\documentclass[letterpaper, 10 pt, conference]{ieeeconf}  

\IEEEoverridecommandlockouts                              

\usepackage[numbers]{natbib}
\usepackage{mathtools}
\usepackage{amsfonts}  
\usepackage{graphics}
\usepackage[pdftex]{graphicx}
\usepackage[font=small]{caption} 

\usepackage{color}
\usepackage{dsfont}
\usepackage[]{mdframed}
\usepackage{algorithm}
\usepackage{algorithmicx}
\usepackage{xparse}
\usepackage{amsmath}
\usepackage{mathtools}
\usepackage{amssymb}
\usepackage{amssymb}
\usepackage{algpseudocode}
\usepackage{bbm}

\usepackage{booktabs}
\usepackage{fancyvrb}

\usepackage{xcolor} 
\usepackage{hyperref} 
\hypersetup{
    colorlinks=true,
}

\usepackage[capitalise]{cleveref}

\DeclarePairedDelimiterX{\infdivx}[2]{(}{)}{%
  #1\;\delimsize\|\;#2%
}

\usepackage{expl3}
\ExplSyntaxOn
\newcommand\latinabbrev[1]{
  \peek_meaning:NTF . {
    #1\@}%
  { \peek_catcode:NTF a {
      #1.\@ }%
    {#1.\@}}}
\ExplSyntaxOff
\def\eg{\latinabbrev{e.g}}

\NewDocumentCommand \proposition {g g g g} {\texttt{#1}(#2
  \IfValueTF{#3}{,\,#3}{}
  \IfValueTF{#4}{,\,#4}{}
  )
}

\NewDocumentCommand \actioncall {g g g g} {\text{#1}(#2
  \IfValueTF{#3}{,#3}{}
  \IfValueTF{#4}{,#4}{}
  \texttt{)}
}

\providecommand{\rebuttal}[1]{{\color{black} #1}}

\definecolor{electricpurple}{rgb}{0.75, 0.0, 1.0}

\def \MethodName {Keypoint Affordance Learning from Imagined Environments}
\def \MethodAcronym {KALIE}

\title{\LARGE \bf
\MethodAcronym: Fine-Tuning Vision-Language Models for \\ Open-World Manipulation without Robot Data
}

\author{Grace Tang$^{*1}$, Swetha Rajkumar$^{*1}$, Yifei Zhou$^{1}$, Homer Rich Walke$^{1}$, Sergey Levine$^{\dagger1}$, Kuan Fang$^{\dagger2}$%
\vspace{0.4cm} \\ 
\url{https://kalie-vlm.github.io/}
\vspace{-0.2cm} 
\thanks{$^*$Equal contribution. $^\dagger$Equal advising.}
\thanks{$^{1}$University of California, Berkeley. $^{2}$Cornell University.}
}

\begin{document}

\maketitle
\thispagestyle{empty}
\pagestyle{empty}

\begin{abstract}

Building generalist robotic systems involves effectively endowing robots with the capabilities to handle novel objects in an open-world setting.
Inspired by the advances of large pre-trained models, we propose \MethodName~(\MethodAcronym), which adapts pre-trained Vision Language Models (VLMs) for robotic control in a scalable manner. Instead of directly producing motor commands, \MethodAcronym~controls the robot by predicting point-based affordance representations based on natural language instructions and visual observations of the scene.
The VLM is trained on 2D images with affordances labeled by humans, bypassing the need for training data collected on robotic systems. Through an affordance-aware data synthesis pipeline, \MethodAcronym~automatically creates massive high-quality training data based on limited example data manually collected by humans. We demonstrate that \MethodAcronym~can learn to robustly solve new manipulation tasks with unseen objects given only 50 example data points. Compared to baselines using pre-trained VLMs, our approach consistently achieves superior performance.

\end{abstract}

\section{Introduction}

The capability to handle an open set of objects, behaviors, and task specifications is essential to the development of generalist robotic systems. Existing learning methods for robotic control can require extensive amounts of data collected on embodied systems~\cite{Levine2016LearningHC, Pinto2015SupersizingSL, brohan2023rt1}.
The diversity and quality of the collected data determines the generalization capability that these methods can achieve, which is subject to robotics expertise and manual labor that humans can provide.
How can we endow robots with generalizable skills for solving an open set of tasks in a scalable manner? 

\begin{figure}[t!]
    \centering
    \includegraphics[width=\columnwidth]{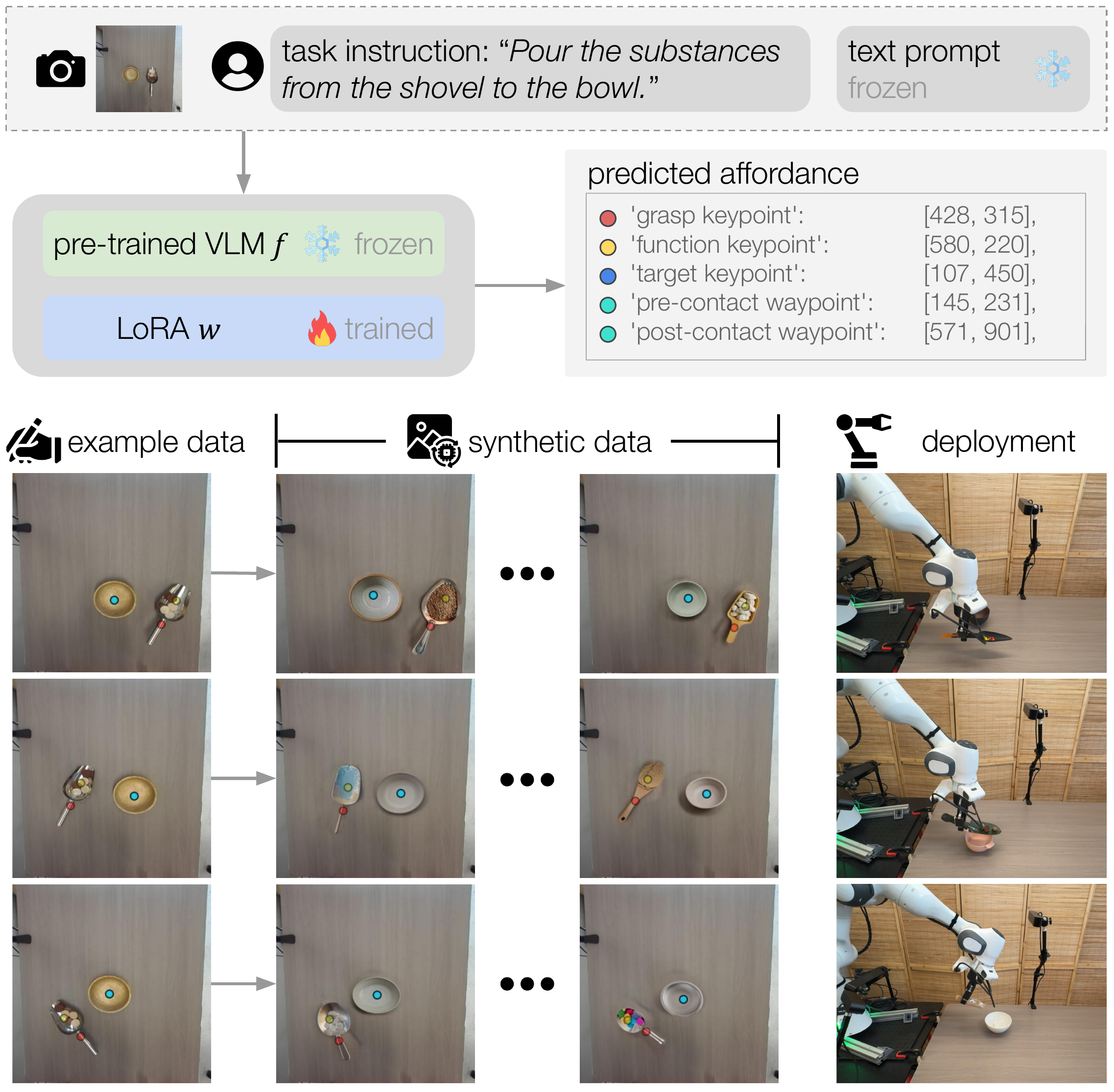}
    \caption{\textbf{Overview of \MethodAcronym}. By fine-tuning a pre-trained VLM, \MethodAcronym~predicts the point-based affordance representation given the input task instruction and visual observation. Based on limited example data collected by humans, \MethodAcronym~generates synthetic data with high diversity while preserving the task semantics and the keypoint annotations. The fine-tuned VLM can robustly generate motions for tasks with unseen objects and arrangements.}
    \label{fig:intro}
\vspace{-5mm}
\end{figure}

Large pre-trained models offer promising tools with their generalist visual understanding and commonsense reasoning abilities~\cite{openai2024gpt4, geminiteam2024gemini, wang2024cogvlm, zhu2023minigpt4}. 
Prior works have shown that pre-trained Large Language Models (LLMs) and Vision Language Models (VLMs) can be directly applied to robotics control through prompt engineering in a zero-shot manner~\cite{liang2023code, huang2023voxposer, fangandliu2024moka, Shah2022LMNavRN}.
However, despite the capability of generalizing to unseen tasks, such systems often suffer from instability and require significant hard-coded domain knowledge to compensate for the pre-trained models' limited knowledge about robotic control. 
While fine-tuning on robotic data can mitigate this issue and proves more sample-efficient than training policies from scratch~\cite{brohan2023rt2, octo_2023}, the largest available datasets~\cite{embodimentcollaboration2024open, khazatsky2024droid, walke2024bridgedata} for robotic control  are still far from comparable to the Internet-scale data used for pre-training the large models with billions of parameters.
It remains a grand challenge to effectively employ and adapt pre-trained large models for robotic control.  

In this paper, we aim to study an alternative solution to this challenge by training large models for manipulation without data collected on robots.
Our key insight is to use visual affordances to guide robotic control and leverage the broad knowledge incorporated in large pre-trained models to efficiently learn to predict the affordances.
Built upon point-based affordance representations defined on 2D images from \citet{Manuelli2019kPAMKA} and \citet{fangandliu2024moka}, we fine-tune a VLM on labeled affordance data.
Humans can easily collect such affordance data by randomizing scenes for a target task (\eg, cleaning a table), taking images through the camera, and annotating the affordances on the image, which bypasses the need for collecting demonstration trajectories through teleoperation of the robots or hard-coded policies.
The main challenges are 1) how to repurpose the VLM pre-trained for visual-question answering for efficient affordance learning, and 2) how to efficiently utilize human supervisions to create training data that can cover diverse scenarios.

To this end, we propose \MethodName~(\MethodAcronym).
As shown in \cref{fig:intro}, our approach fine-tunes a pre-trained VLM to a point-based affordance representation given the input task instruction and visual observation. 
Starting with limited example data of manually arranged scenes and annotated affordance labels, \MethodAcronym~automatically creates a massive number of diverse synthetic images to scale up the training data.
To stay faithful to the task semantics and keypoint annotations while diversifying the data distribution as much as possible, we propose an affordance-aware data synthesis pipeline using a pre-trained diffusion model~\cite{rombach2022highresolution, zhang2023adding} with additional contexts.

We evaluate \MethodAcronym~on various manipulation tasks involving tool-use, articulated objects, and deformable objects. 
\MethodAcronym~consistently solves target tasks with diverse unseen objects and initial arrangements and achieves superior performances compared to baselines using pre-trained VLMs~\cite{huang2023voxposer, fangandliu2024moka}.
\section{Related Work}

\subsection{Robotic Control with Large Models}

An increasing number of works have employed foundation models in robotics through pre-training or fine-tuning on robot data manually collected through teleoperation or scripted policies~\cite{brohan2023rt2, Driess2023PaLMEAE, octo_2023, shridhar2022cliport}.
Due to the lack of datasets that can cover the vast complexity and diversity of robotics applications, most of the approaches focus on specific task categories such as grasping and object rearrangements.
The generalization capability of the trained model is also constrained to the distribution of objects and environments covered by the manually collected datasets.  
Alternatively, other works have attempted to combine and prompt pre-trained models to solve unseen tasks in zero-shot manners~\cite{liang2023code, huang2023voxposer, fangandliu2024moka, Shah2022LMNavRN, hu2023look}.
However, the performance of these works is usually subject to the capability of the pre-trained models, as well as non-trivial expert knowledge and manual labor to design prompts and in-context examples.  
In contrast to these approaches, the proposed method fine-tunes a VLM to robustly solve the target tasks. 
To avoid the need of collecting extensive amount of robot data, our model trains the VLM to predict the point-based affordance representation from \cite{fangandliu2024moka} and employs a diffusion model to automatically synthesize massive, high-quality data.

\subsection{Adaptation of Vision Language Models} 

Due to their versatile nature, Vision Language Models (VLMs) can be effectively fine-tuned to accommodate a variety of downstream tasks~\citep{hong2023cogagent, zhai2024finetuning, chen2023fireact, zeng2023agenttuning}. Specifically, to facilitate the prediction of spatially grounded outputs, previous research has investigated methods such as sets of marks~\citep{yang2023setofmark}, scaffolding~\citep{openai2024gpt4}, and coordinate-based bounding boxes~\citep{wang2023onepeace, chen2023shikra, wang2024cogvlm}. Of these methods, coordinate-based references are particularly adaptable, capable of pinpointing any location within an image. Thus, we employ this approach for predicting keypoints. Our contribution does not lie in proposing a new method for fine-tuning VLMs, but rather in integrating and adapting VLM fine-tuning for robotic control. 

\subsection{Data Synthesis for Robot Learning} 

To alleviate the data bottleneck in robot learning, prior works have investigated various approaches to augment and synthesize data. 
Data augmentation, especially random image transformations, have been widely used to improve generalization to unseen visual inputs~\cite{Kostrikov2020ImageAI, laskin2020reinforcement}.
These random operations can effectively improve the models generalization capabilities to unseen visual inputs during test time, but cannot extend the model's capabilities beyond the coverage of the training distribution.
Domain randomization has also been broadly used to train robust models for robotic control~\cite{tobin2017domain, mehta2020active, akkaya2019solving, qi2023general} when collecting simulated robot experiences.
In contrast, our method directly diversifies the training data without the need for a physical simulator.
Recent works in computer vision and robotics have leveraged deep generative models, such as diffusion models, to synthesize unseen environments by leveraging broad knowledge learned from Internet-scale images~\cite{kar2019meta, zhang2021datasetgan, wang2023robogen, yang2023learning, Yu2023ScalingRL, devaranjan2020meta, wood2021fake, kim2023neuralfield, li2022bigdatasetgan}.
While the prior work can easily generate defected samples, we propose an affordance-aware data synthesis approach that enables the diffusion model to generate diverse data with much higher quality and consistent affordance annotations. 
\section{Keypoint Affordance Learning from \\ Imagined Environments}


We propose \MethodName~(\MethodAcronym) to adapt pre-trained Vision Language Models (VLMs) to acquire generalizable skills without robot experiences.
In this section, we will first define the affordance prediction problem in the few-shot setting using point-based affordances labeled by human experts.
Next, we will introduce a novel affordance-aware data synthesis recipe to diversify the training data, which automatically generates massive high-quality data based on the example data collected only for limited scenarios.
Then, we will describe our VLM fine-tuning approach and discuss the key design options.
Lastly, we summarize the overall system at the end of this section.

\subsection{Problem Statement}

We consider the problem of open-world robotic manipulation involving unseen objects and initial arrangements of the scenes.
As shown in \cref{fig:intro}, each task is single-stage and is specified by a free-form language description $l$, such as ``\textit{Use the brush to sweep the snack package.}''
The robot observes an RGBD image from a third-person camera and performs a 6-DoF motion trajectory in open loop to complete the task.

To tackle this problem, we employ a VLM $f$ pre-trained on Internet-scale data~\cite{wang2024cogvlm}. 
Following the practice of \citep{fangandliu2024moka}, we query the VLM to produce point-based affordance representations to guide a motion planner to generate motions.
The VLM takes as inputs the prompt $\rho$, the task instruction $l$, and the input image $s$ and predict the affordance representation $y$ as:
\begin{equation}
    \hat{y} = f(\rho, l , s),
\end{equation}
where $y$ contains a set of keypoints, including the grasp keypoint, the function keypoint, the target keypoint, the pre-contact waypoint, and the post-contact waypoint, specified by 2D coordinates on the image (see \cref{fig:intro}).
Additional properties such as the height and the orientation of the gripper will also be decided for each task. 
Based on the affordance representation, a low-level motion generator computes a motion trajectory to complete the task. 
We assume the desired motion to solve each task can be specified by the same subset of these points (\eg, the sweeping with a brush requires all five points, and drawer closing requires everything but the grasp point, as shown in \cref{sec:qualitative_results}), but the specific coordinates of these points depend on the objects and their poses with respect to the robot.

In this work, we consider tasks and objects which are challenging for VLMs to handle in zero-shot.
In contrast to \cite{fangandliu2024moka}, we consider a few-shot learning setting, in which we fine-tune the pre-trained VLM to acquire and improve skills for unseen scenarios on limited and non-robot data.
Notably, the point-based representations enable us to outsource motion generation to the low-level motion planner and focus on only the affordance prediction problem. 
Therefore, training only requires data of pairs of an observed image $s$ and the ground truth keypoints $y$ collected by human experts or generated automatically as described below.  
We assume access to an example dataset $\mathcal{D}$ containing a limited number of $(s, y)$ pairs for each target task. 
In our experiments, we assume the most extreme case, in which the data is collected on \textit{a single set of} objects for each task and the fine-tuned model is evaluated on unseen object sets. 

\subsection{Affordance-Aware Data Synthesis}
\label{sec:data_synthesis}

To scale up the training of the VLM to enhance its generalization capabilities to unseen scenarios, we automatically synthesize a training dataset $\mathcal{D}'$ to cover a wide range of environments.
Each of the new data points $(s', y') \in \mathcal{D}'$ is synthesized by modifying an existing data point $(s, y) \in \mathcal{D}$ collected by human experts.

Following the practice of \cite{Yu2023ScalingRL}, we compute the segmentation mask of the object in the scene using open-vocabulary segmentation~\cite{Kirillov2023SegmentA} and then inpaint the masked region with a diffusion model~\cite{rombach2022highresolution}.
However, naively inpainting the masked region can lead to undesired results.
Without a direct mechanism to specify the geometric properties of the object, it would be hard to diversify the inpainted images in a way that can cover the desired distribution of testing scenarios.
Moreover, there is usually a discrepancy between the appearance of the inpainted object and the original keypoint annotation, introducing the need to manually re-label keypoints.
To generate massive, high-quality data without additional manual labor, we need to ensure that the generated images stay faithful to the context of the target task and the annotated keypoints, while aggressively diversifying the environments as much as possible.

\begin{figure}[t!]
    \centering
    \includegraphics[width=\linewidth]{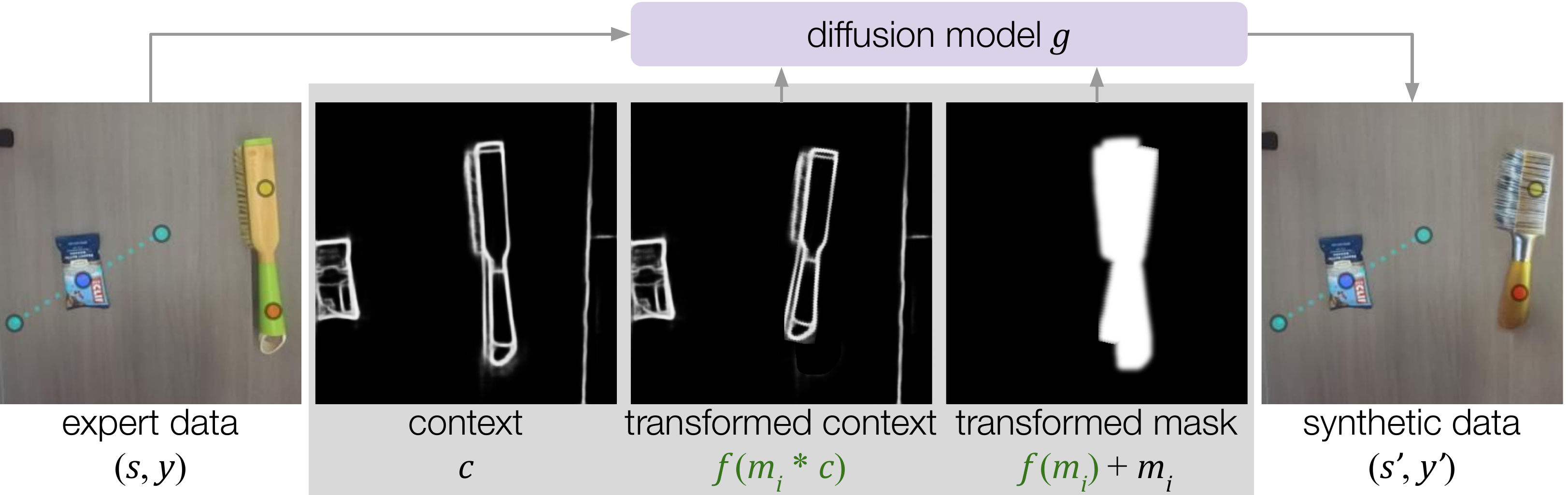}
    \caption{\textbf{Affordance-aware data synthesis.} \MethodAcronym~employs the inpainting capability of a diffusion model to generate synthetic data. To diversify the scenes while staying faithful to the task semantics and the keypoint annotations, \MethodAcronym~computes and transforms the context, such as soft edges, to guide the generation process.}
    \label{fig:data_synthesis}
    \vspace{-5mm}
\end{figure}

To tackle this challenge, we design an affordance-aware data synthesis pipeline by leveraging additional context images to guide the generation process, as shown in \cref{fig:data_synthesis}.
Specifically, we employ the ControlNet~\cite{zhang2023adding} diffusion model $g$, which takes inputs as the input image $s$, the segmentation mask $m$, a context image $c$, and a language description of the object $o$ and generates the new image as $s' \sim g(\cdot | s, m, c, o)$.
We would like $c$ to be a compact representation of the object's geometric properties that provide clues about the affordances, while minimizing other detailed visual information to leave enough free reign for the diffusion model.
By inpainting an image $s'$ in accordance with $c$, we hope to obtain new objects of the same point-based affordances.
In this work, we choose to use a soft edge map as $c$ computed by an external image processing algorithm~\cite{Soria_2023teed}, which outlines the contours and parts of the object. 

To cover testing objects of unseen shapes, we introduce additional randomization operations to the geometry of the synthesized object. 
Directly modifying either the object's appearance in the pixel space can easily affect other parts of the image and create artifacts.  
Instead, we propose to apply transformation $h(\cdot)$ on the compact context $c$ before calling the diffusion model $G$ to inpaint the image. 
The transformation function $h(\cdot)$ can include basic data augmentation operations such as random scaling, translation, and rotation, as well as additional operations such as elastic distortion.
To modify the context $c$, we transform the region under the masked as $h(m * c)$.
The region to be inpainted on the original image now becomes $m + h(m)$, which includes the transformed silhouette of the object $h(m)$ in addition to the white space left by removing the original object $m$. 
Accordingly, we also apply the same transformation to the annotated keypoints $y$ with the context image to keep them consistent.
Therefore, the sampling process with the transformation can be written as:
\begin{eqnarray}
    s' & \sim & g(\cdot | s, h(m) + m, h(m * c), o), \\
    y' & = & h(y), \nonumber
\end{eqnarray}
where we slightly overload the notation by using $h(\cdot)$ to denote the transformation applied to both the images and the keypoint coordinates.  

\subsection{Efficient Adaptation for Keypoint Affordance Prediction}
\label{sec:vlm_finetuning}

We fine-tune the VLM $f$ to predict point-based affordance representations.
To adapt $f$, which is pre-trained for visual-question answering by predicting tokens, we need to make our design choices around two considerations. 
First, how to represent the point-based affordances so that we can effectively re-purpose or modify the VLM's prediction head for affordance prediction.  
Second, how to perform sample-efficient fine-tuning to utilize the pre-trained VLM's pre-trained capabilities while endowing it with the additional knowledge from the new dataset $\mathcal{D}$. 

We investigate two design options for the prediction head.
\textbf{Regression Head}~\citep{chen2024visionlanguage} adds an additional linear layer on top of the last hidden state of the VLM to directly predict the $x-y$ coordinates of the keypoint affordances. 
The input to the VLM is the image along with the appropriate task instructions, and the last hidden state of the last token is used.
\textbf{Natural Language Affordance Prediction} fine-tunes the VLM to output a well-formatted natural language that includes text-based keypoint affordances, with an example shown in \cref{fig:intro}.
Each keypoint affordance is represented by the $x-y$ coordinates as integers normalized between a predefined range. 

These design options resort to the pre-trained weights and the new dataset in different manners. Empirically as shown in \cref{fig:ablation}, we found that the two design choices achieve similar performances. As Natural Language Affordance Prediction aligns more closely with other applications of VLMs such as Visual Question Answering, we choose to use this design option as our main method. During training, we convert the ground truth affordance label $y$ into the corresponding format to compute the losses.
For both design options, we use Low-Rank Adaptation (LoRA)~\cite{liu2023grounding} to fine-tune the VLM. We use an L2 regression loss for Regression Head and the cross entropy loss for Natural Language Affordance Prediction.

\subsection{System Summary}

The overall pipeline of \MethodAcronym~is summarized in \cref{algo:imagination}. 
Starting with the original example dataset $\mathcal{D}$, \MethodAcronym~automatically creates $\mathcal{D}'$ to fine-tune the VLM $f$.
The data synthesis pipeline operates in an object-centric manner, iteratively processing and inpainting each of the $M$ objects in the scene as explained in \cref{sec:data_synthesis}, using each object's segmentation mask $m_i$ and description $o_i$.
During this process, a VLM (which can be the same as $f$ or a different model) is used to sample an alternative description of the object to guide the inpainting process of the diffusion model $g$.  
The generated data is combined with the example dataset to adapt $f$ by optimizing the weights $w$ as explained in \cref{sec:vlm_finetuning}.

\begin{algorithm}
\caption{\MethodName~(\MethodAcronym)}
\textbf{Inputs}: Pre-trained VLM $f$, task instruction $l$, example dataset $D$, desired dataset size $N$.
\begin{algorithmic}[1]

\State $D' \gets \varnothing$
\While{$|D'| < N$}
    \State Sample $(s, y)$ from $D$.
    \State Compute the context image $c$ given $s$.
    \State $s'_0 \gets s$
    \For{ $i = 1, ..., M$}  (~\cref{sec:data_synthesis})
        \State Extract the description $o_i$ given $s$ and $l$.
        \State Get the segmentation mask $m_i$ given $s$ and $o_i$.
        \State Sample a new description $o'_i$ given $o_i$.
        \State Transform the mask, context as $h(m_i)$, $h(m_i * c)$.
        \State Transform the keypoints as $h(y_i)$.
        \State Inpaint $s'_i \sim G(\cdot | s'_{i - 1}, h(m_i) + m_i, h(m_i * c), o'_i)$.
    \EndFor
    \State Merge the transformed keypoints into $y'$.
\EndWhile
\State Fine-tune $f$ on $D \cup D'$ by optimizing the weights $w$ (~\cref{sec:vlm_finetuning}).
\end{algorithmic}
\label{algo:imagination}
\end{algorithm}
\section{Experiments}

We design our experiments to investigate the following questions: 1) Can \MethodAcronym~synthesize diverse and high-quality data for affordance prediction? 2) Can \MethodAcronym~fine-tune pre-trained VLMs to improve their performances on challenging manipulation tasks with unseen objects? 
3) What design options are critical to the performance of \MethodAcronym?

\subsection{Experimental Setup}

\textbf{Environments and tasks.} 
Our experiments are conducted in a real-world table-top manipulation environment with a 7-DoF robot arm and a Robotiq 2F-85 gripper.
A top-down RGBD camera is used to receive visual observation of the environment. 
We design 5 table-top manipulation tasks (table sweeping, drawer closing, towel hanging, trowel pouring, and usb unplugging) each with 3 different testing object sets unseen during training.

\textbf{Pre-trained models.}  We fine-tune CogVLM-17B~\cite{wang2024cogvlm} using
LoRA~\cite{liu2023grounding} with 6 layers and rank 10. The data synthesis is conducted using GPT-4V~\cite{openai2024gpt4}, which has higher capacity but does not have open-sourced code and models.

\textbf{Affordance representations and motion primitive.} \MethodAcronym~follows the keypoint-based affordance representation defined in \cite{fangandliu2024moka}, which contains the grasp keypoint, function keypoint, target keypoint, pre-contact waypoint, and post-contact waypoint. 
Given the affordance representation, the motion trajectory is produced by the motion generator in the SE(3) space, which consists of a grasping phase and a manpulation phase~\cite{fangandliu2024moka}.

\textbf{Training and evaluation.} 
To fine-tune the VLM, 50 example images are collected by humans through randomly setting up the scenes based on the context of the task and annotating the affordance representations with a graphical user interface. 
Through data synthesis with diffusion models, we generate an additional set of 500 pairs of images and keypoints to the 50 examples. 
We fine-tune the VLM using the Adam optimizer~\cite{kingma2017adammethodstochasticoptimization} with a weight decay of $5\times10^{-2}$ and apply an annealing learning rate scheduler with a starting learning rate of $1\times10^{-5}$ and a cosine learning rate style. 
We train our models for 6000 iterations using a batch size of 4.
For each target task, we test the model on 15 trials with 3 sets of unseen objects. Empirically, fine-tuning takes around 8 hours on a single Nvidia A100 SXM 80GB GPU. 

\textbf{Data scalability.} 
Collecting each image and annotating the keypoints takes a few seconds, depending on the task complexity, which is significantly faster and easier than collecting a robot trajectory as done in prior work on the same experimental setup~\cite{khazatsky2024droid}. Empirically, collecting and generating the 500 synthetic images takes 90 minutes in total on a single Nvidia A100 SXM 80GB GPU.

\subsection{Data Synthesis}

\begin{figure}[t!]
    \centering
    \includegraphics[width=\linewidth]{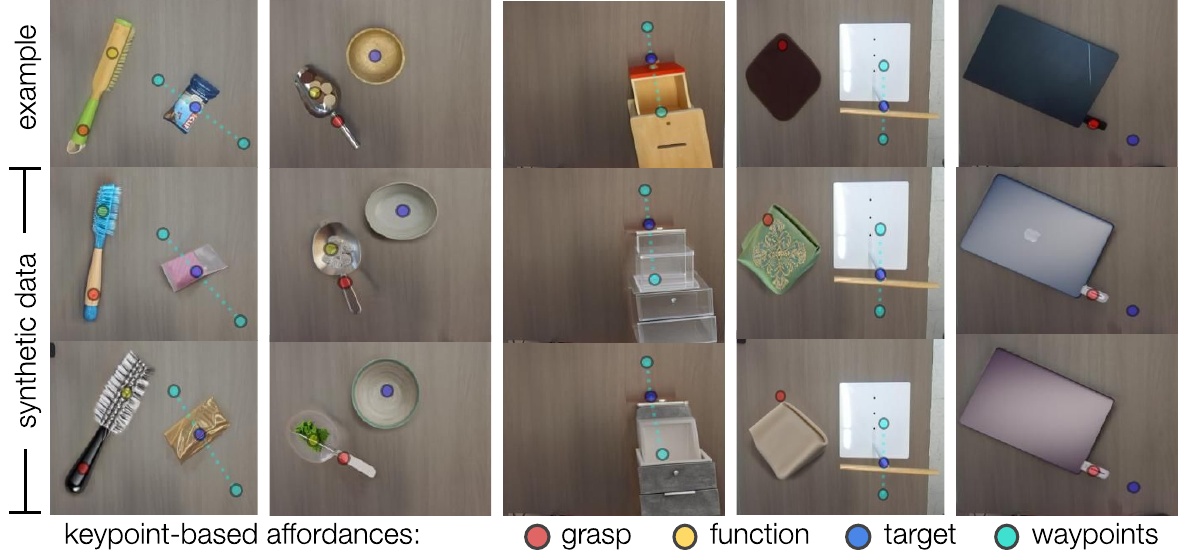}
    \caption{\textbf{Synthetic data examples. }
    In each column, we show example synthetic images generated based on an example image for each task.
The original and transformed point-based affordances are plotted on top of the images.
    \vspace{-5mm}
    }
    \label{fig:imagination_examples}
\end{figure}

In \cref{fig:imagination_examples}, we show examples of the synthetic images. Each row contains the data for a task. The first row is the example data and the remaining rows are the synthetic data. \MethodAcronym~  demonstrates generating diverse samples while preserving the task context and the original affordance labels. The generated objects have diverse, realistic textures that are different from the original image, including the transparent surfaces of the drawers and the detailed bristles on the brushes. Most of the examples are indistinguishable from the real example data with only minor artifacts at the boundary of the masked original image. 

\begin{figure}[t!]
    \centering
    \includegraphics[width=\columnwidth]{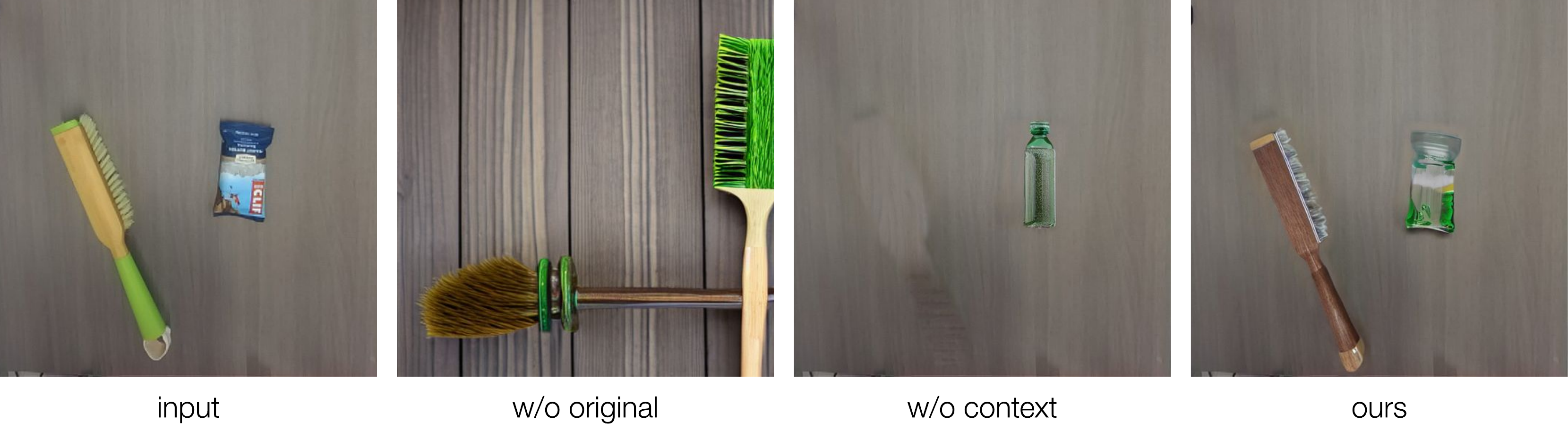}
    \caption{\textbf{Comparisons with alternative synthesis algorithms.} \MethodAcronym~generates much more robust samples comparing with generation without conditioning on the original images or the context.
    }
    \vspace{-3mm}
    \label{fig:generation_comp}
\end{figure}

To demonstrate the difficulty of such synthesis problems and the importance of the design options in \MethodAcronym, we also analyze the results from alternative approaches as seen in \Cref{fig:generation_comp}.
We first directly apply the diffusion model to generate the image without grounding on the original example image. Not only is the generation irrelevant to the annotated keypoints, without further guidance the quantity and appearance of the generated objects are also misaligned with the input instruction. 
While the original ROSIE~\cite{Yu2023ScalingRL} does not have an open-source codebase and uses an internal diffusion model, we reimplement it with the non-contextual version of the same diffusion model~\cite{zhang2023adding} as in \MethodAcronym.
Without the context information, the model ignores part of the prompt and inpaints an empty table surface for the brush and a completely different shape for the trash. 
We found that this issue is particularly common for more complex objects, such as tools, but less severe for simpler items like towels, as shown in \cite{Yu2023ScalingRL} and our experiments.

\begin{table}[tb!]
    \centering
    \setlength{\tabcolsep}{4pt}
    \begin{tabular}{lcccccc}  
    \toprule
    & \multicolumn{1}{c}{Table} & \multicolumn{1}{c}{Drawer} & \multicolumn{1}{c}{Towel} & \multicolumn{1}{c}{Trowel}  & \multicolumn{1}{c}{USB}  \\
    Methods & \multicolumn{1}{c}{Sweeping} & \multicolumn{1}{c}{Closing} & \multicolumn{1}{c}{Hanging} & \multicolumn{1}{c}{Pouring}  & \multicolumn{1}{c}{Unplugging}  \\
    \midrule
    VoxPoser~\citep{huang2023voxposer} & 3/15 & 8/15 &  1/15 & 0/15 & 0/15    \\
    MOKA~\citep{fangandliu2024moka} & 9/15 & 9/15 & 5/15  & 7/15 & 2/15     \\
    \MethodAcronym~(Ours) & \textbf{14}/15 & \textbf{15}/15 & \textbf{13}/15 & \textbf{13}/15 & \textbf{9}/15  \\
    \bottomrule
    \end{tabular}
    \caption{\textbf{Task success rates.} We evaluate \MethodAcronym~on five manipulation tasks involving tool use, deformable objects, and articulated objects. \MethodAcronym~robustly solves these tasks and consistently achieves superior performances compared to baselines. }
    \label{tab:experiment}
    \vspace{-5mm}
\end{table}

\subsection{Comparative Results}

We compare \MethodAcronym~with two baselines that employ VLMs for robotic manipulation, \textbf{VoxPoser}~\citep{huang2023voxposer} and \textbf{MOKA}~\citep{fangandliu2024moka}. In contrast to \MethodAcronym, which fine-tunes a much smaller open-source VLM, the two baselines use the pre-trained OpenAI GPT-4V\cite{openai2024gpt4}.
We use the collected example data as the in-context examples for MOKA.

We evaluate each method across 5 manipulation tasks involving tool objects, articulated objects, and deformable objects.
Compared to the baseline methods using pre-trained VLMs, \MethodAcronym~consistently achieves higher success rates.
For comparison, MOKA~\cite{fangandliu2024moka} is subject to failures such as incorrect output formats and affordance reasoning, while Voxposer~\cite{huang2023voxposer}'s failure cases are often due to perception errors and a lack of in-context examples for more complex tasks, such as tool use.
The VLM fine-tuned by \MethodAcronym~robustly solves the tasks without these issues, and has far fewer failures due to inaccurate point predictions.  

\subsection{Ablative Study}

\begin{figure}[tb!]
    \centering
    \includegraphics[width=\linewidth]{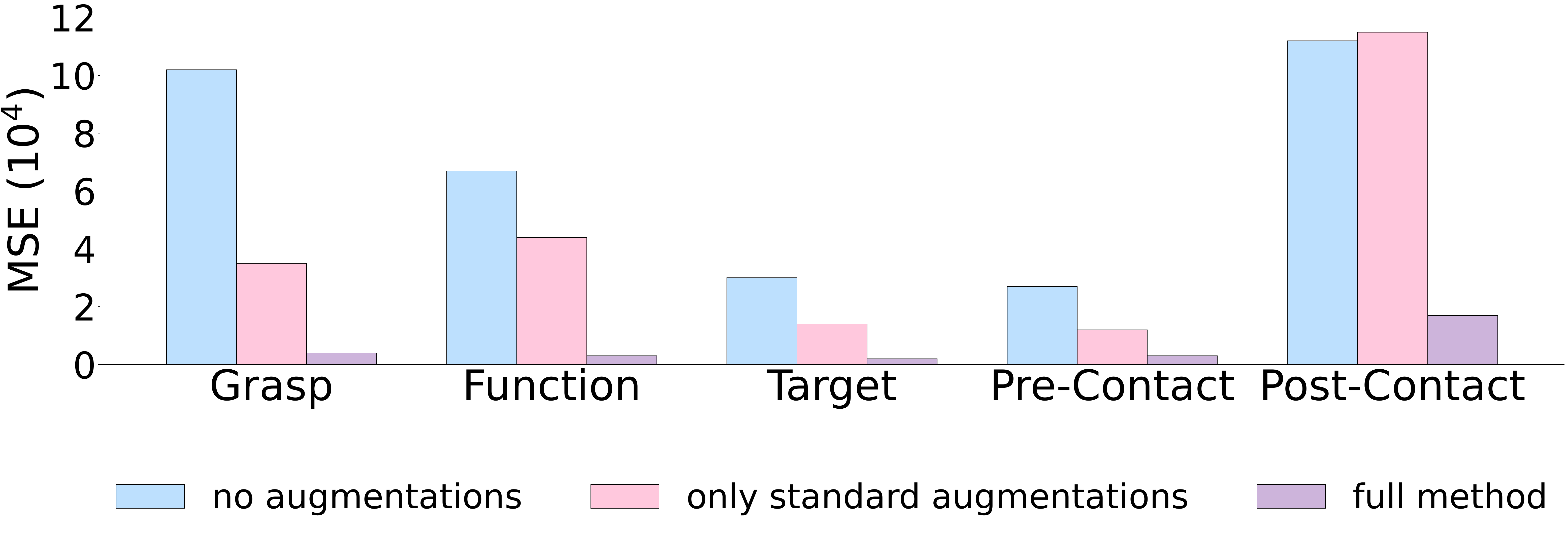}
    \caption{\textbf{Comparisons with vanilla data augmentation.} 
    Mean Square Error (MSE) for each keypoint affordance on a test set of novel objects is reported for the table sweeping task.
    }
    \vspace{-3mm}
    \label{fig:augmentations}
\end{figure}

\begin{figure}[tb!]
    \centering
    \includegraphics[width=\linewidth]{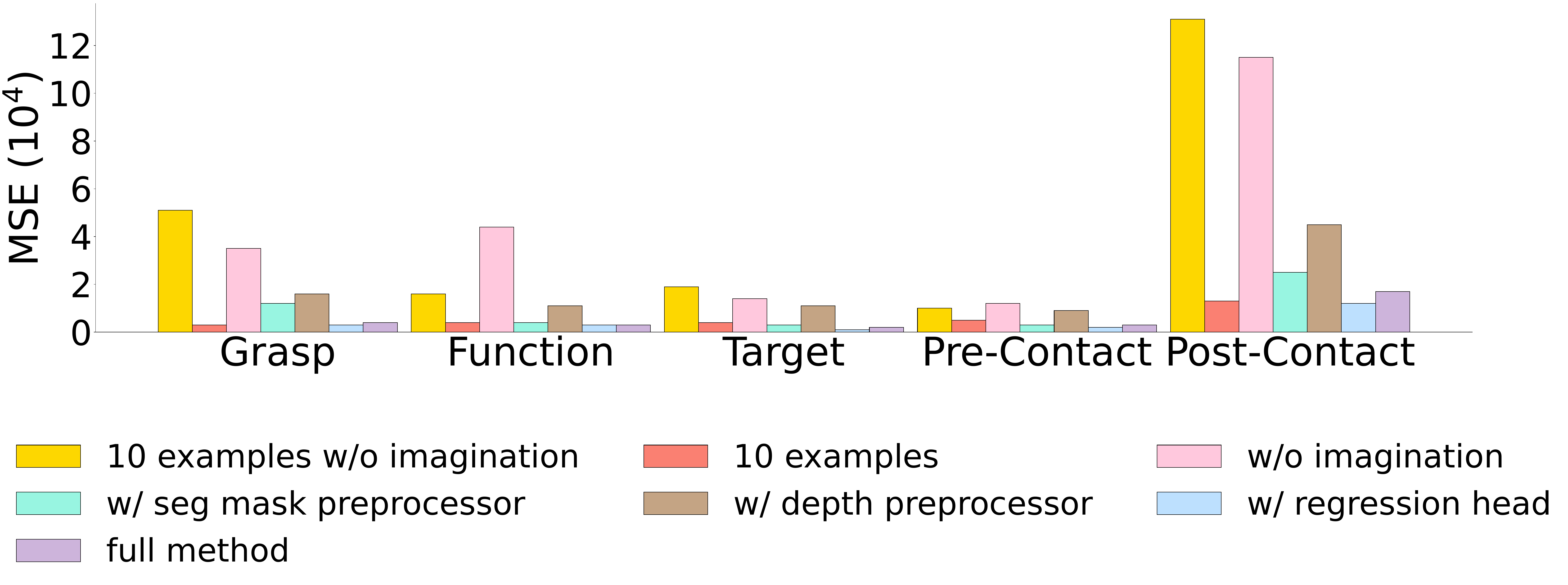}
    \caption{\rebuttal{\textbf{Ablative study.} The ablation experiments are carried out on the sweeping task. Mean Square Error (MSE) for each keypoint affordance on a test set of novel objects is reported.}}
    \label{fig:ablation}
    \vspace{-5mm}
\end{figure}

We conduct an in-depth ablation study to study the performance of \MethodAcronym~as seen in \Cref{fig:augmentations} and \Cref{fig:ablation}. Detailed analysis and discussions are provided below.

\textbf{Comparisons with vanilla data augmentation.}
In \cref{fig:augmentations}, we quantify the impact of both vanilla image augmentations and the synthetic data by \MethodAcronym~on the VLM's performance. The standard set of image augmentations includes rotations, random resized crops, horizontal/vertical flipping, and color jittering, which are added in the main training loop. 
We compare the performance of our model fine-tuned using \MethodAcronym~with vanilla augmentation (\textbf{full method}) to that only using vanilla augmentations (\textbf{only standard augmentations}) and without any augmentations (\textbf{no augmentations}). 
We report the test mean square error (MSE) on a held-out set of 50 real images with unseen objects. 
We observe that using both sets of augmentations yields the best results in the VLM's prediction of keypoint affordances. 

\textbf{Comparisons of prediction heads.} In \cref{fig:ablation}, we analyze the choice of Natural Language Affordance Prediction \rebuttal{(\textbf{full method})} against Regression Head \rebuttal{(\textbf{w/ regression head})}. We report the MSE on a held-out set with unseen objects and ground-truth annotations. The coordinates are normalized between 0 and 999 with respect to the actual height and width of the image. The train set consists of 50 real images collected on the robot and 500 synthesized images, and the test set consists of 50 real images of novel objects excluded from the train set. We observe that Imagined Environments significantly improve the test MSE and our method with Natural Language Affordance Prediction works comparably with Regression Head, while staying consistent with the language interface of VLMs for vision-language tasks.

\textbf{Comparisons with alternative context types. } We examine two alternative types of context information, depth mapping and segmentation masking, to provide ControlNet with the context image. In \cref{fig:ablation} we include MSE values on the same test set when using both these preprocessors, which we refer to as \textbf{w/ depth preprocessor} and \textbf{w/ seg mask preprocessor}, respectively. We found that both the depth map and the segmentation mask preprocessor models have inferior performances. We hypothesize that while these preprocessors yield the correct general shape of the objects, they do not capture enough detail for ControlNet~\cite{zhang2023adding} to be effective. Conversely, the softedge preprocessing allows for a generalization of the shape of each object while maintaining sufficient details of the object shape.

\textbf{Scalability with example data.} We further investigated the results of training on 550 generated data points from 10 real images, as well as training a model on only the 10 real images. We refer to these as \textbf{10 examples} and \textbf{10 examples w/o imagination} respectively. As shown in \cref{fig:ablation}, while 10 examples w/o imagination clearly performs worse than our other methods, we do not see any major difference between the MSEs of the 10 examples and 50 examples (full method). This further shows that our method is easily scalable, as a dataset with only 10 expert-annotated examples is able to perform at the level of a dataset with 50 such examples. 

\begin{figure}[t!]
    \centering
    \includegraphics[width=\columnwidth]{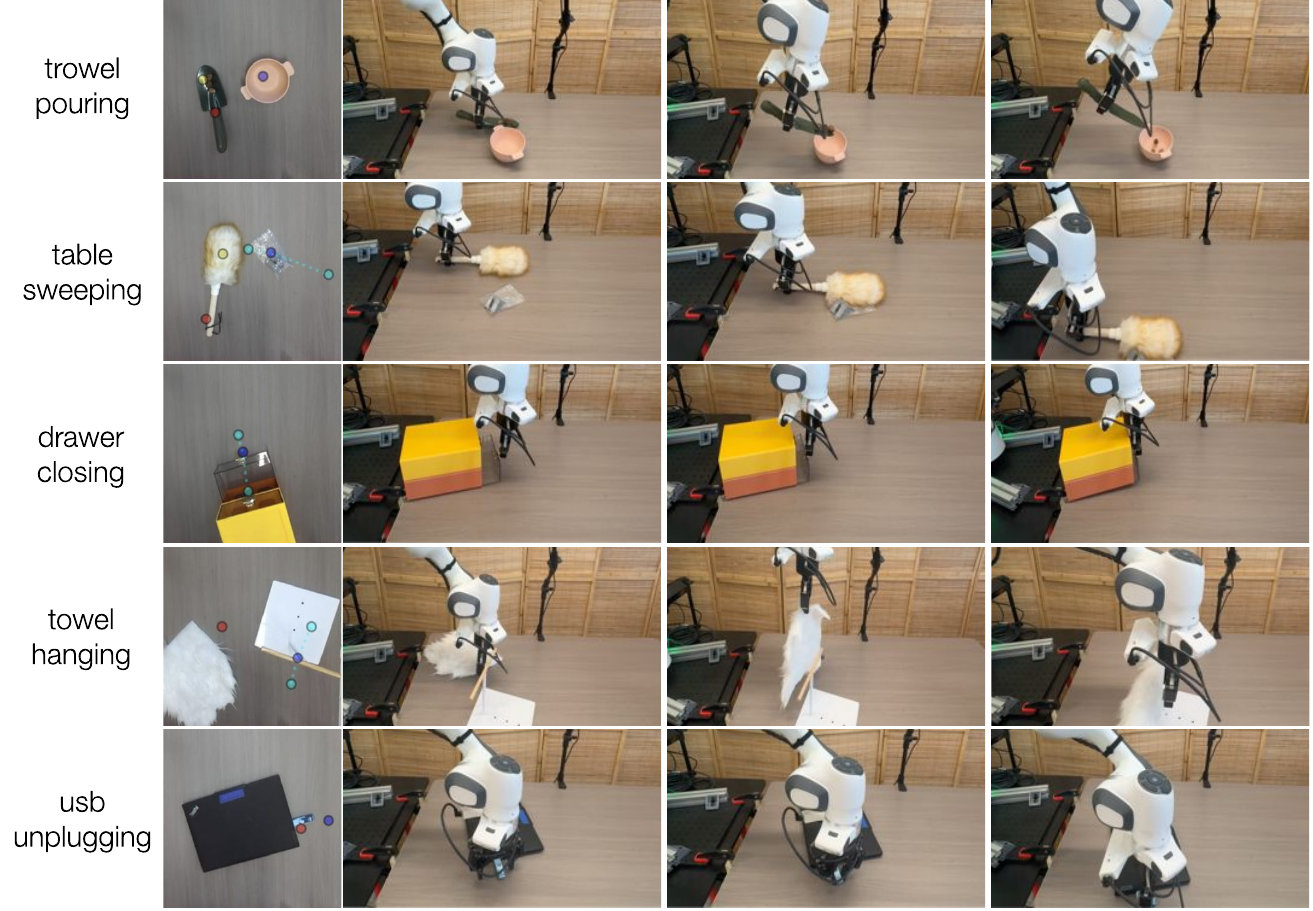}
    \caption{\textbf{Affordance prediction and task execution. } We show examples of the predicted affordance representations and the key frames of task execution. Across 5 manipulation tasks, each with 3 sets of unseen objects, \MethodAcronym~robustly solves the task.}
    \vspace{-5mm}
    \label{fig:task_execution}
\end{figure}

\subsection{Task Execution}
\label{sec:qualitative_results}

We show examples of affordance predictions and task execution in \cref{fig:task_execution}. The VLM fine-tuned by \MethodAcronym~successfully predicts the affordance and solves each of the tasks. As shown in the figure, the robot successfully solve the tasks through interactions with the environment given the predicted keypoint-based affordance representations.
\section{Conclusion}

In this work, we propose \MethodAcronym~to fine-tune pre-trained VLMs to predict affordances for robotic manipulation with open sets of objects and initial arrangements of scenes. 
Using the proposed affordance-aware data synthesis pipeline, we generate massive high-quality data to scale up training without extensive manual labor or domain expertise with robots. 
We fine-tune the VLM on the generated data to predict the point-based affordance representations. 
Across various manipulation tasks involving tool use, deformable objects, and articulated objects, we demonstrate \MethodAcronym~can robustly solve the task and consistently outperform baselines. 
We hope \MethodAcronym~will inspire future research towards adapting vision-language models for open-world robotic control.

\textbf{Limitations and future work.} 
The current affordance representation in \MethodAcronym~is still limited to single-arm table-top manipulation. 
To apply \MethodAcronym~to more complicated scenarios, such as dynamic manipulation and whole-body control, we would need to extend the design of the affordance representation. 
In addition, there is still a large discrepancy between the open-source VLM fine-tuned by \MethodAcronym~and the state-of-the-art VLMs without accessible fine-tuning APIs.
Lastly, we have only discussed few-shot generalization in this paper, and would need to train KALIE on more diverse set of tasks to generalize to unseen target tasks in a zero-shot manner.
In the future, we hope to apply \MethodAcronym~to more powerful VLMs to further improve its performance.

\section*{ACKNOWLEDGMENT}
This research was supported by the AI Institute, AFOSR FA9550-22-1-0273, and the NSF under IIS-2246811. 
We would like to thank Sudeep Dasari, Ria Doshi, Stefanie Gschwind, Fangchen Liu, Cyprien Noel, Karl Pertsch, and Paul Zhou for valuable supports with the infrastructure.


\bibliographystyle{plainnat} 
\bibliography{references}

\begin{thebibliography}{49}
\providecommand{\natexlab}[1]{#1}
\providecommand{\url}[1]{\texttt{#1}}
\expandafter\ifx\csname urlstyle\endcsname\relax
  \providecommand{\doi}[1]{doi: #1}\else
  \providecommand{\doi}{doi: \begingroup \urlstyle{rm}\Url}\fi

\bibitem[Akkaya et~al.(2019)Akkaya, Andrychowicz, Chociej, Litwin, McGrew, Petron, Paino, Plappert, Powell, Ribas, et~al.]{akkaya2019solving}
Ilge Akkaya, Marcin Andrychowicz, Maciek Chociej, Mateusz Litwin, Bob McGrew, Arthur Petron, Alex Paino, Matthias Plappert, Glenn Powell, Raphael Ribas, et~al.
\newblock Solving rubik's cube with a robot hand.
\newblock \emph{arXiv preprint arXiv:1910.07113}, 2019.

\bibitem[Brohan et~al.(2023{\natexlab{a}})]{brohan2023rt1}
Anthony Brohan et~al.
\newblock Rt-1: Robotics transformer for real-world control at scale, 2023{\natexlab{a}}.

\bibitem[Brohan et~al.(2023{\natexlab{b}})]{brohan2023rt2}
Anthony Brohan et~al.
\newblock Rt-2: Vision-language-action models transfer web knowledge to robotic control, 2023{\natexlab{b}}.

\bibitem[Chen et~al.(2023{\natexlab{a}})Chen, Shu, Shareghi, Collier, Narasimhan, and Yao]{chen2023fireact}
Baian Chen, Chang Shu, Ehsan Shareghi, Nigel Collier, Karthik Narasimhan, and Shunyu Yao.
\newblock Fireact: Toward language agent fine-tuning, 2023{\natexlab{a}}.

\bibitem[Chen et~al.(2023{\natexlab{b}})Chen, Zhang, Zeng, Zhang, Zhu, and Zhao]{chen2023shikra}
Keqin Chen, Zhao Zhang, Weili Zeng, Richong Zhang, Feng Zhu, and Rui Zhao.
\newblock Shikra: Unleashing multimodal llm's referential dialogue magic, 2023{\natexlab{b}}.

\bibitem[Chen et~al.(2024)Chen, Mees, Kumar, and Levine]{chen2024visionlanguage}
William Chen, Oier Mees, Aviral Kumar, and Sergey Levine.
\newblock Vision-language models provide promptable representations for reinforcement learning, 2024.

\bibitem[Collaboration(2024)]{embodimentcollaboration2024open}
Embodiment Collaboration.
\newblock Open x-embodiment: Robotic learning datasets and rt-x models, 2024.

\bibitem[Devaranjan et~al.(2020)Devaranjan, Kar, and Fidler]{devaranjan2020meta}
Jeevan Devaranjan, Amlan Kar, and Sanja Fidler.
\newblock Meta-sim2: Unsupervised learning of scene structure for synthetic data generation.
\newblock In \emph{Computer Vision--ECCV 2020: 16th European Conference, Glasgow, UK, August 23--28, 2020, Proceedings, Part XVII 16}, pages 715--733. Springer, 2020.

\bibitem[Driess et~al.(2023)Driess, Xia, Sajjadi, Lynch, Chowdhery, Ichter, Wahid, Tompson, Vuong, Yu, Huang, Chebotar, Sermanet, Duckworth, Levine, Vanhoucke, Hausman, Toussaint, Greff, Zeng, Mordatch, and Florence]{Driess2023PaLMEAE}
Danny Driess, F.~Xia, Mehdi S.~M. Sajjadi, Corey Lynch, Aakanksha Chowdhery, Brian Ichter, Ayzaan Wahid, Jonathan Tompson, Quan~Ho Vuong, Tianhe Yu, Wenlong Huang, Yevgen Chebotar, Pierre Sermanet, Daniel Duckworth, Sergey Levine, Vincent Vanhoucke, Karol Hausman, Marc Toussaint, Klaus Greff, Andy Zeng, Igor Mordatch, and Peter~R. Florence.
\newblock Palm-e: An embodied multimodal language model.
\newblock In \emph{International Conference on Machine Learning}, 2023.
\newblock URL \url{https://api.semanticscholar.org/CorpusID:257364842}.

\bibitem[Fang et~al.(2024)Fang, Liu, Abbeel, and Levine]{fangandliu2024moka}
Kuan Fang, Fangchen Liu, Pieter Abbeel, and Sergey Levine.
\newblock Moka: Open-world robotic manipulation through mark-based visual prompting.
\newblock 2024.

\bibitem[Hong et~al.(2023)Hong, Wang, Lv, Xu, Yu, Ji, Wang, Wang, Zhang, Li, Xu, Dong, Ding, and Tang]{hong2023cogagent}
Wenyi Hong, Weihan Wang, Qingsong Lv, Jiazheng Xu, Wenmeng Yu, Junhui Ji, Yan Wang, Zihan Wang, Yuxuan Zhang, Juanzi Li, Bin Xu, Yuxiao Dong, Ming Ding, and Jie Tang.
\newblock Cogagent: A visual language model for gui agents, 2023.

\bibitem[Hu et~al.(2023)Hu, Lin, Zhang, Yi, and Gao]{hu2023look}
Yingdong Hu, Fanqi Lin, Tong Zhang, Li~Yi, and Yang Gao.
\newblock Look before you leap: Unveiling the power of gpt-4v in robotic vision-language planning.
\newblock \emph{arXiv preprint arXiv:2311.17842}, 2023.

\bibitem[Huang et~al.(2023)Huang, Wang, Zhang, Li, Wu, and Fei-Fei]{huang2023voxposer}
Wenlong Huang, Chen Wang, Ruohan Zhang, Yunzhu Li, Jiajun Wu, and Li~Fei-Fei.
\newblock Voxposer: Composable 3d value maps for robotic manipulation with language models.
\newblock \emph{arXiv preprint arXiv:2307.05973}, 2023.

\bibitem[Kar et~al.(2019)Kar, Prakash, Liu, Cameracci, Yuan, Rusiniak, Acuna, Torralba, and Fidler]{kar2019meta}
Amlan Kar, Aayush Prakash, Ming-Yu Liu, Eric Cameracci, Justin Yuan, Matt Rusiniak, David Acuna, Antonio Torralba, and Sanja Fidler.
\newblock Meta-sim: Learning to generate synthetic datasets.
\newblock In \emph{Proceedings of the IEEE/CVF International Conference on Computer Vision}, pages 4551--4560, 2019.

\bibitem[Khazatsky et~al.(2024)]{khazatsky2024droid}
Alexander Khazatsky et~al.
\newblock Droid: A large-scale in-the-wild robot manipulation dataset, 2024.

\bibitem[Kim et~al.(2023)Kim, Brown, Yin, Kreis, Schwarz, Li, Rombach, Torralba, and Fidler]{kim2023neuralfield}
Seung~Wook Kim, Bradley Brown, Kangxue Yin, Karsten Kreis, Katja Schwarz, Daiqing Li, Robin Rombach, Antonio Torralba, and Sanja Fidler.
\newblock Neuralfield-ldm: Scene generation with hierarchical latent diffusion models.
\newblock In \emph{Proceedings of the IEEE/CVF Conference on Computer Vision and Pattern Recognition}, pages 8496--8506, 2023.

\bibitem[Kingma and Ba(2017)]{kingma2017adammethodstochasticoptimization}
Diederik~P. Kingma and Jimmy Ba.
\newblock Adam: A method for stochastic optimization, 2017.

\bibitem[Kirillov et~al.(2023)Kirillov, Mintun, Ravi, Mao, Rolland, Gustafson, Xiao, Whitehead, Berg, Lo, Doll{\'a}r, and Girshick]{Kirillov2023SegmentA}
Alexander Kirillov, Eric Mintun, Nikhila Ravi, Hanzi Mao, Chloe Rolland, Laura Gustafson, Tete Xiao, Spencer Whitehead, Alexander~C. Berg, Wan-Yen Lo, Piotr Doll{\'a}r, and Ross~B. Girshick.
\newblock Segment anything.
\newblock \emph{2023 IEEE/CVF International Conference on Computer Vision (ICCV)}, pages 3992--4003, 2023.
\newblock URL \url{https://api.semanticscholar.org/CorpusID:257952310}.

\bibitem[Kostrikov et~al.(2020)Kostrikov, Yarats, and Fergus]{Kostrikov2020ImageAI}
Ilya Kostrikov, Denis Yarats, and Rob Fergus.
\newblock Image augmentation is all you need: Regularizing deep reinforcement learning from pixels.
\newblock \emph{ArXiv}, abs/2004.13649, 2020.
\newblock URL \url{https://api.semanticscholar.org/CorpusID:216562627}.

\bibitem[Laskin et~al.(2020)Laskin, Lee, Stooke, Pinto, Abbeel, and Srinivas]{laskin2020reinforcement}
Misha Laskin, Kimin Lee, Adam Stooke, Lerrel Pinto, Pieter Abbeel, and Aravind Srinivas.
\newblock Reinforcement learning with augmented data.
\newblock \emph{Advances in neural information processing systems}, 33:\penalty0 19884--19895, 2020.

\bibitem[Levine et~al.(2016)Levine, Pastor, Krizhevsky, and Quillen]{Levine2016LearningHC}
Sergey Levine, Peter Pastor, Alex Krizhevsky, and Deirdre Quillen.
\newblock Learning hand-eye coordination for robotic grasping with deep learning and large-scale data collection.
\newblock \emph{The International Journal of Robotics Research}, 37:\penalty0 421 -- 436, 2016.

\bibitem[Li et~al.(2022)Li, Ling, Kim, Kreis, Fidler, and Torralba]{li2022bigdatasetgan}
Daiqing Li, Huan Ling, Seung~Wook Kim, Karsten Kreis, Sanja Fidler, and Antonio Torralba.
\newblock Bigdatasetgan: Synthesizing imagenet with pixel-wise annotations.
\newblock In \emph{Proceedings of the IEEE/CVF Conference on Computer Vision and Pattern Recognition}, pages 21330--21340, 2022.

\bibitem[Liang et~al.(2023)Liang, Huang, Xia, Xu, Hausman, Ichter, Florence, and Zeng]{liang2023code}
Jacky Liang, Wenlong Huang, Fei Xia, Peng Xu, Karol Hausman, Brian Ichter, Pete Florence, and Andy Zeng.
\newblock Code as policies: Language model programs for embodied control.
\newblock In \emph{IEEE International Conference on Robotics and Automation}, pages 9493--9500. IEEE, 2023.

\bibitem[Liu et~al.(2023)Liu, Zeng, Ren, Li, Zhang, Yang, Li, Yang, Su, Zhu, and Zhang]{liu2023grounding}
Shilong Liu, Zhaoyang Zeng, Tianhe Ren, Feng Li, Hao Zhang, Jie Yang, Chunyuan Li, Jianwei Yang, Hang Su, Jun Zhu, and Lei Zhang.
\newblock Grounding dino: Marrying dino with grounded pre-training for open-set object detection, 2023.

\bibitem[Manuelli et~al.(2019)Manuelli, Gao, Florence, and Tedrake]{Manuelli2019kPAMKA}
Lucas Manuelli, Wei Gao, Peter~R. Florence, and Russ Tedrake.
\newblock kpam: Keypoint affordances for category-level robotic manipulation.
\newblock In \emph{International Symposium of Robotics Research}, 2019.
\newblock URL \url{https://api.semanticscholar.org/CorpusID:80628296}.

\bibitem[Mehta et~al.(2020)Mehta, Diaz, Golemo, Pal, and Paull]{mehta2020active}
Bhairav Mehta, Manfred Diaz, Florian Golemo, Christopher~J Pal, and Liam Paull.
\newblock Active domain randomization.
\newblock In \emph{Conference on Robot Learning}, pages 1162--1176. PMLR, 2020.

\bibitem[{Octo Model Team} et~al.(2023){Octo Model Team}, Ghosh, Walke, Pertsch, Black, Mees, Dasari, Hejna, Xu, Luo, Kreiman, Tan, Sadigh, Finn, and Levine]{octo_2023}
{Octo Model Team}, Dibya Ghosh, Homer Walke, Karl Pertsch, Kevin Black, Oier Mees, Sudeep Dasari, Joey Hejna, Charles Xu, Jianlan Luo, Tobias Kreiman, {You Liang} Tan, Dorsa Sadigh, Chelsea Finn, and Sergey Levine.
\newblock Octo: An open-source generalist robot policy.
\newblock \url{https://octo-models.github.io}, 2023.

\bibitem[OpenAI(2024)]{openai2024gpt4}
OpenAI.
\newblock Gpt-4 technical report, 2024.

\bibitem[Pinto and Gupta(2015)]{Pinto2015SupersizingSL}
Lerrel Pinto and Abhinav~Kumar Gupta.
\newblock Supersizing self-supervision: Learning to grasp from 50k tries and 700 robot hours.
\newblock \emph{2016 IEEE International Conference on Robotics and Automation (ICRA)}, pages 3406--3413, 2015.

\bibitem[Qi et~al.(2023)Qi, Yi, Suresh, Lambeta, Ma, Calandra, and Malik]{qi2023general}
Haozhi Qi, Brent Yi, Sudharshan Suresh, Mike Lambeta, Yi~Ma, Roberto Calandra, and Jitendra Malik.
\newblock General in-hand object rotation with vision and touch.
\newblock In \emph{Conference on Robot Learning}, pages 2549--2564. PMLR, 2023.

\bibitem[Rombach et~al.(2022)Rombach, Blattmann, Lorenz, Esser, and Ommer]{rombach2022highresolution}
Robin Rombach, Andreas Blattmann, Dominik Lorenz, Patrick Esser, and Björn Ommer.
\newblock High-resolution image synthesis with latent diffusion models, 2022.

\bibitem[Shah et~al.(2022)Shah, Osinski, Ichter, and Levine]{Shah2022LMNavRN}
Dhruv Shah, Blazej Osinski, Brian Ichter, and Sergey Levine.
\newblock Lm-nav: Robotic navigation with large pre-trained models of language, vision, and action.
\newblock In \emph{Conference on Robot Learning}, 2022.

\bibitem[Shridhar et~al.(2022)Shridhar, Manuelli, and Fox]{shridhar2022cliport}
Mohit Shridhar, Lucas Manuelli, and Dieter Fox.
\newblock Cliport: What and where pathways for robotic manipulation.
\newblock In \emph{Conference on Robot Learning}, pages 894--906. PMLR, 2022.

\bibitem[Soria et~al.(2023)Soria, Li, Rouhani, and Sappa]{Soria_2023teed}
Xavier Soria, Yachuan Li, Mohammad Rouhani, and Angel~D. Sappa.
\newblock Tiny and efficient model for the edge detection generalization.
\newblock In \emph{Proceedings of the IEEE/CVF International Conference on Computer Vision (ICCV) Workshops}, pages 1364--1373, October 2023.

\bibitem[Team(2024)]{geminiteam2024gemini}
Gemini Team.
\newblock Gemini: A family of highly capable multimodal models, 2024.

\bibitem[Tobin et~al.(2017)Tobin, Fong, Ray, Schneider, Zaremba, and Abbeel]{tobin2017domain}
Josh Tobin, Rachel Fong, Alex Ray, Jonas Schneider, Wojciech Zaremba, and Pieter Abbeel.
\newblock Domain randomization for transferring deep neural networks from simulation to the real world.
\newblock In \emph{2017 IEEE/RSJ international conference on intelligent robots and systems (IROS)}, pages 23--30. IEEE, 2017.

\bibitem[Walke et~al.(2024)Walke, Black, Lee, Kim, Du, Zheng, Zhao, Hansen-Estruch, Vuong, He, Myers, Fang, Finn, and Levine]{walke2024bridgedata}
Homer Walke, Kevin Black, Abraham Lee, Moo~Jin Kim, Max Du, Chongyi Zheng, Tony Zhao, Philippe Hansen-Estruch, Quan Vuong, Andre He, Vivek Myers, Kuan Fang, Chelsea Finn, and Sergey Levine.
\newblock Bridgedata v2: A dataset for robot learning at scale, 2024.

\bibitem[Wang et~al.(2023{\natexlab{a}})Wang, Wang, Lin, Bai, Zhou, Zhou, Wang, and Zhou]{wang2023onepeace}
Peng Wang, Shijie Wang, Junyang Lin, Shuai Bai, Xiaohuan Zhou, Jingren Zhou, Xinggang Wang, and Chang Zhou.
\newblock One-peace: Exploring one general representation model toward unlimited modalities, 2023{\natexlab{a}}.

\bibitem[Wang et~al.(2024)Wang, Lv, Yu, Hong, Qi, Wang, Ji, Yang, Zhao, Song, Xu, Xu, Li, Dong, Ding, and Tang]{wang2024cogvlm}
Weihan Wang, Qingsong Lv, Wenmeng Yu, Wenyi Hong, Ji~Qi, Yan Wang, Junhui Ji, Zhuoyi Yang, Lei Zhao, Xixuan Song, Jiazheng Xu, Bin Xu, Juanzi Li, Yuxiao Dong, Ming Ding, and Jie Tang.
\newblock Cogvlm: Visual expert for pretrained language models, 2024.

\bibitem[Wang et~al.(2023{\natexlab{b}})Wang, Xian, Chen, Wang, Wang, Fragkiadaki, Erickson, Held, and Gan]{wang2023robogen}
Yufei Wang, Zhou Xian, Feng Chen, Tsun-Hsuan Wang, Yian Wang, Katerina Fragkiadaki, Zackory Erickson, David Held, and Chuang Gan.
\newblock Robogen: Towards unleashing infinite data for automated robot learning via generative simulation.
\newblock \emph{arXiv preprint arXiv:2311.01455}, 2023{\natexlab{b}}.

\bibitem[Wood et~al.(2021)Wood, Baltru{\v{s}}aitis, Hewitt, Dziadzio, Cashman, and Shotton]{wood2021fake}
Erroll Wood, Tadas Baltru{\v{s}}aitis, Charlie Hewitt, Sebastian Dziadzio, Thomas~J Cashman, and Jamie Shotton.
\newblock Fake it till you make it: face analysis in the wild using synthetic data alone.
\newblock In \emph{Proceedings of the IEEE/CVF international conference on computer vision}, pages 3681--3691, 2021.

\bibitem[Yang et~al.(2023{\natexlab{a}})Yang, Zhang, Li, Zou, Li, and Gao]{yang2023setofmark}
Jianwei Yang, Hao Zhang, Feng Li, Xueyan Zou, Chunyuan Li, and Jianfeng Gao.
\newblock Set-of-mark prompting unleashes extraordinary visual grounding in gpt-4v.
\newblock \emph{arXiv preprint arXiv:2310.11441}, 2023{\natexlab{a}}.

\bibitem[Yang et~al.(2023{\natexlab{b}})Yang, Du, Ghasemipour, Tompson, Schuurmans, and Abbeel]{yang2023learning}
Mengjiao Yang, Yilun Du, Kamyar Ghasemipour, Jonathan Tompson, Dale Schuurmans, and Pieter Abbeel.
\newblock Learning interactive real-world simulators.
\newblock \emph{arXiv preprint arXiv:2310.06114}, 2023{\natexlab{b}}.

\bibitem[Yu et~al.(2023)Yu, Xiao, Stone, Tompson, Brohan, Wang, Singh, Tan, Dee, Peralta, Ichter, Hausman, and Xia]{Yu2023ScalingRL}
Tianhe Yu, Ted Xiao, Austin Stone, Jonathan Tompson, Anthony Brohan, Su~Wang, Jaspiar Singh, Clayton Tan, Meredith Dee, Jodilyn Peralta, Brian Ichter, Karol Hausman, and F.~Xia.
\newblock Scaling robot learning with semantically imagined experience.
\newblock \emph{ArXiv}, abs/2302.11550, 2023.

\bibitem[Zeng et~al.(2023)Zeng, Liu, Lu, Wang, Liu, Dong, and Tang]{zeng2023agenttuning}
Aohan Zeng, Mingdao Liu, Rui Lu, Bowen Wang, Xiao Liu, Yuxiao Dong, and Jie Tang.
\newblock Agenttuning: Enabling generalized agent abilities for llms, 2023.

\bibitem[Zhai et~al.(2024)Zhai, Bai, Lin, Pan, Tong, Zhou, Suhr, Xie, LeCun, Ma, and Levine]{zhai2024finetuning}
Yuexiang Zhai, Hao Bai, Zipeng Lin, Jiayi Pan, Shengbang Tong, Yifei Zhou, Alane Suhr, Saining Xie, Yann LeCun, Yi~Ma, and Sergey Levine.
\newblock Fine-tuning large vision-language models as decision-making agents via reinforcement learning, 2024.

\bibitem[Zhang et~al.(2023)Zhang, Rao, and Agrawala]{zhang2023adding}
Lvmin Zhang, Anyi Rao, and Maneesh Agrawala.
\newblock Adding conditional control to text-to-image diffusion models, 2023.

\bibitem[Zhang et~al.(2021)Zhang, Ling, Gao, Yin, Lafleche, Barriuso, Torralba, and Fidler]{zhang2021datasetgan}
Yuxuan Zhang, Huan Ling, Jun Gao, Kangxue Yin, Jean-Francois Lafleche, Adela Barriuso, Antonio Torralba, and Sanja Fidler.
\newblock Datasetgan: Efficient labeled data factory with minimal human effort.
\newblock In \emph{Proceedings of the IEEE/CVF Conference on Computer Vision and Pattern Recognition}, pages 10145--10155, 2021.

\bibitem[Zhu et~al.(2023)Zhu, Chen, Shen, Li, and Elhoseiny]{zhu2023minigpt4}
Deyao Zhu, Jun Chen, Xiaoqian Shen, Xiang Li, and Mohamed Elhoseiny.
\newblock Minigpt-4: Enhancing vision-language understanding with advanced large language models, 2023.

\end{thebibliography}

\end{document}